# Geolocation differences of language use in urban areas

Olga Kellert[1] and Nicholas H. Matlis[2]

[1]*Georg-August-Universität Göttingen*

[2]*Center for Free-Electron Laser Science, Deutsches Elektronen Synchrotron, Notkestraße 85, 22607 Hamburg, Germany.*

**Abstract**.

The explosion in the availability of natural language data in the era of social media has given rise to a host of applications such as sentiment analysis and opinion mining. Simultaneously, the growing availability of precise geolocation information is enabling visualization of global phenomena such as environmental changes and disease propagation. Opportunities for tracking spatial variations in language use, however, have largely been overlooked, especially on small spatial scales. Here we explore the use of Twitter data with precise geolocation information to resolve spatial variations in language use on an urban scale down to single city blocks. We identify several categories of language tokens likely to show distinctive patterns of use and develop quantitative methods to visualize the spatial distributions associated with these patterns. Our analysis concentrates on comparison of contrasting pairs of Tweet distributions from the same category, each defined by a set of tokens. Our work shows that analysis of small-scale variations can provide unique information on correlations between language use and social context which are highly valuable to a wide range of fields from linguistic science and commercial advertising to social services.

1. Introduction

Studying how language use varies from location to location has long been a key tool for linguistic analysis (see Hovy et al 2020 and references therein). These studies can provide important clues to a range of issues, from tracking the origins of specific phenomena to understanding the mechanisms of language change generally. Until recently, most linguistic studies have suffered from the lack of large data sets, the unavailability of precise location information and the use of manual processing methods (Labov 2000, Smakman & Heinrich 2017, among others). The developments of technologies such as the global positioning system (GPS) and social media platforms such as Twitter have the potential to revolutionize linguistic research by providing it with orders of magnitude more information and precision. The advantages of tracking geolocation information have already been exploited for studying a variety of natural and social phenomena such as virus propagation, environmental pollution and car-crash reports (Su et al 2020, Milusheva et al 2021, among others). These studies have provided critical information and visualization tools for tracking rapid changes and spatial variations of social and natural events. Use of geolocation information for linguistic research, however, is very limited, leaving unexplored a major opportunity for development of new linguistic methods. The few studies that do exist have mostly concentrated on large spatial scales corresponding to cities and countries (henceforth "macro-scale") (e.g., Gonçalves & Sánchez 2016, Sánchez et al 2018, Hovy et al 2020, Grieve et al 2019, Eisenstein et al 2014, Nguyen et al 2016, Izbicki et al 2019). To our knowledge, no work has so far been reported which examines geolocation differences of linguistic variation on smaller spatial scales like urban areas (henceforth "micro-scale").

Studying geolocation of linguistic variation on the *micro-scale* opens up the possibility of correlating language use with social context and social groups. Establishing such correlations enables geolocated social media text to be used for efficient collection of social-context information which is otherwise difficult to obtain and is highly valuable for analyzing socially-relevant issues, including politics, commerce, social welfare and linguistics. In commerce, for instance, analysis of product-related keywords could allow more effective placement of advertising. Similarly, analysis of school-related words could provide information on variations in the quality of education. For linguistics, this tool has the potential to transform our understanding of language-variation and language-change mechanisms by providing direct clues as to the origins of these variations. A major breakthrough would be to track down the origins of emerging language variations to particular locations within a city and observe their propagation over time. These few examples just hint at the large range of possible applications.

Here we take the first steps in exploring the feasibility of studying language variations on the *micro-scale*. In particular, we demonstrate that it is indeed possible to spatially resolve language variations on scales down to a single city block. As this analysis is new, the key question to answer in determining its usefulness is whether the pattern of observed variations can be reliably correlated to an underlying phenomenon. In order to address this question, we examine a series of cases designed to quantify the variation originating from several sources, including random statistical variations, variations in linguistic elements directly correlated to the local features of the city, and finally variations in the use of dialects. We also implement statistical tests to estimate the significance of these results.

The outline of the paper is as follows. Section 2 describes the methodology and Section 3 shows the results of the applied methodology. Section 4 discusses the findings and outlines the new contribution of this paper and suggests further studies in the future research.

## 2. Methodology

In this section, we describe the rationale behind the choice of our data set and describe the methodology used to calculate and compare distributions of Tweets satisfying various criteria.

### 2.1 Data

The data was sourced from the Twitter database because it is freely accessible and because geolocation information is included. The geolocation information is included in a subset of Tweets in the form of both GPS coordinates associated with the city as well as the precise location from which the Tweets were sent. For our analysis, we used only those Tweets with the precise coordinates, which correspond to somewhat less than 1% of the total number of Tweets which contain geolocation information.

Using Tweet data as a corpus for scientific research brings many advantages. The most important of these are the large volumes of data and the presence of metadata which can be used to track attributes of the data such as its geolocation, the language used and the time at which it was written. These advantages come at the cost of known biases associated with the social role of the Twitter platform as well as due to large-scale automation of Tweets by "bots," among other things (see Nguyen et al 2016, and reference therein). It is therefore a non-trivial question whether Twitter data or social media in general is suitable or reliable for use in language-based analyses. This issue is a large and highly complex one which we do not attempt to resolve here. Instead we outline several key challenges associated with use of Twitter data.

Despite the fact that nearly 10,000 Tweets are currently tweeted per second (https://www.internetlivestats.com/), data scarcity remains one of the most significant issues. As expected, the number of Tweets available for analysis reduces as the size of the regions studied becomes smaller. Therefore, performing analysis on the *micro-scale* places higher demands on the data than for the *macro-scale*. This issue becomes more pronounced as the topic of the analysis becomes more specific. The range of topics to which this analysis can be applied is thus biased towards those with higher representation in the Corpus. To find sufficient data for *micro-scale* analysis, cities are therefore a natural choice. In cities, the concentration of social media users can be extremely high, providing large amounts of data even within a single city block and opening up the possibility of exploring language variation on the *micro-scale*. The concentration of social media users within a city is by no means uniform, however, necessitating care in comparing one region to another.

To demonstrate our method, we focused on analysis on Ciudad Autónoma de Buenos Aires (CABA) because of its high density of Twitter use and because of its rich multicultural background, due to imigration in the late 18th and early 19th century (Conde 2011, among others). This intermixing of cultures creates an ideal situation to study language variation. The precise area considered is defined by the '*extent*', which corresponds to a longitude range of -58.531725 to -58.355148, and a latitude range of -34.538162 to -34.705446, which includes CABA. The Tweets were further filtered to include only those written in Spanish (i.e., Lang = 'es'), and the time-span of the available data included late 2017 up to early 2021. The result was a set of 693.450 Tweets, which we refer to as the *Corpus*.

We performed our analysis on several distinct cases in order to highlight different categories of language use. For each case, two contrasting sets of Tweets, belonging to the same category, were selected and then compared. Tweets were selected for each set by checking the text for the presence of one or more "tokens" which were manually chosen, based on their linguistic function, to fit a specific category. Here we use *tokens* to refer to various types of linguistic elements, such as words, letters, phrases or even emoticons found in the text of the Tweets.

## 2.2 Binning and visualization on maps

The basis of our computational approach is to calculate the spatial distribution of a set of Tweets defined by an associated set of tokens. The spatial distribution quantifies how the density of Tweets varies within the *extent* chosen for the city. By density, we mean the number of Tweets, satisfying the selection criterion, counted within a fixed area determined by the bin size. The analysis of language variation is then done by comparing the shape of the distributions associated with two sets of Tweets selected to have a contrasting relationship. For clarity, we refer to one of these two sets as the "target" and the other as the "reference". To calculate the distribution, we define a 2-dimensional grid of bins distributed over the spatial extent of the city and count the number of Tweets belonging to each bin. We divide the spatial extent into 200 x 200 bins in order to resolve variations occurring on a scale of roughly 1 km or smaller. This level of resolution allows distinction of neighborhoods and regions within the city corresponding to various functionalities, such as business and residence.

In the following paragraph, we describe our analytical formalism for calculating and comparing the distributions. The spatial extent is defined by the minimum and maximum longitudes, $X_{min}$, $X_{max}$, respectively and the minimum and maximum latitudes, $Y_{min}$, $Y_{max}$, respectively. The longitudinal and latitudinal sizes of the bins are then given by $\Delta x = (X_{max} - X_{min})/n$ and $\Delta y = (Y_{max} - Y_{min})/m$, respectively, where $n = 200$ and $m = 200$ are the number of longitude and latitude bins, respectively. It should be noted that this approach creates bins with uniform size with respect to the longitude and latitude, but varying size with respect to physical distance. This nonuniformity is due to the fact that lines of constant longitude are farthest apart at the equator and become nearer to each other as one approaches the North or South poles. This effect, however, is negligible on the scale of a city, so we can assume the bins, as defined above, encompass a uniform physical area.

For every bin in the grid, we count the number of tweets, $c^A_{i,j}$, whose GPS coordinates lie within the boundaries of the bin in the $i^{th}$ column and $j^{th}$ row of the grid. The set of points $c^A_{i,j}$ represents the spatial distribution of all Tweets tweeted within the city. The target distribution is then represented by the quantity, $c^T_{i,j}$, which is the count of Tweets in the (i,j)th bin which contain at least one of a target set of tokens, while $c^R_{i,j}$ represents the reference distribution and is the count of Tweets in the (i,j)th bin which contain at least one of a reference set of tokens.

The calculation of the distributions was implemented in Python (see Python), and in order to visualize the distributions, markers representing the Tweet counts from each bin were overlaid onto a geographical map using the Cartopy package (see Cartopy). Figure 1 is a plot of $c^A_{i,j}$.

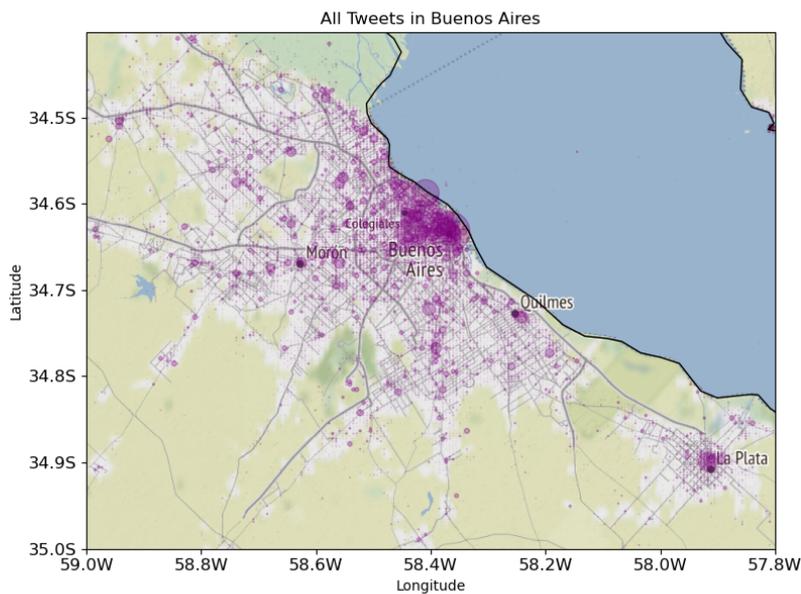

**Figure 1.** Geographical plot of all Tweets in CABA

This figure shows the distribution of all Tweets from our Corpus. In this plot, the number of Tweets in each bin is represented by the size of the marker which are semi-transparent to avoid obscuration of neighboring markers. Overlapping of multiple markers deepens the color, helping to emphasize locations with large Tweet densities.

Inspection of the distribution shows that the Tweet density tracks the regions of development of the city and hence the population density. Specifically, the Tweet density is highest in the city center and follows the major roadways. This general pattern is to be expected for all Tweet distributions, since, all other things being equal, more people create more Tweets. As a result, distinguishing variations between target and reference distributions can be challenging, especially when the variations are small. Figure 2 shows an example where the target and reference are two language varieties spoken in Buenos Aires: Argentinian Spanish and Peninsular Spanish. The two distributions appear nearly identical. A sensitive comparison of distributions therefore requires a representation which emphasizes locations where the distributions vary. In linguistic analysis, a frequently-used metric to compare elements of two categories is the count ratio. This approach, however, can be problematic for comparing Tweet distributions because of the large density differences between the city center and its edges. Since many bins end up with only a few, or even zero Tweets, dividing one distribution by the other then leads to large amounts of noise for bins that have few Tweets and division errors for those with zero Tweets. Some researchers have solved this issue by defining bins of variable size to ensure that the Tweet counts in each are comparable (Schlosser et al 2021, among others). This approach, however, can lead to a loss of spatial resolution, as bins in low-density areas can become very large. Another approach is to only look at those bins where the count is above a defined threshold. In this case, however, the results can vary significantly with the choice of threshold, requiring justification of a particular value. Here we pursue a different approach which preserves spatial resolution but avoids the division-by-zero and thresholding issues: we calculate the mathematical difference between the two Tweet-count distributions.

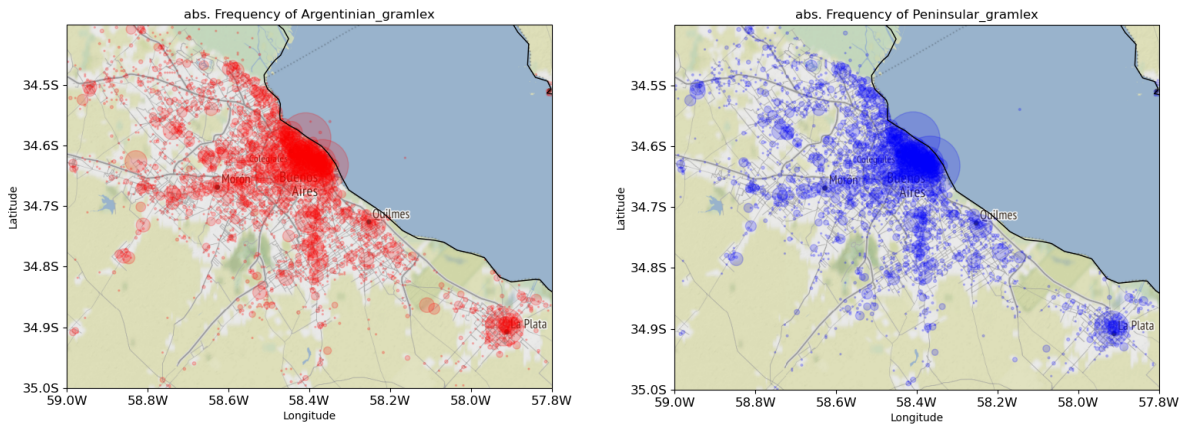

**Figure 2. Left:** Distribution of ArgSp Tweets. **Right:** Distribution of PenSp Tweets.

Since the target and reference distributions generally contain a different total number of Tweets, subtracting one distribution from the other requires first normalizing by the number of Tweets in each. The number of Tweets in the target and reference distributions are expressed as: $N_T \equiv \sum_{i=0}^{n-1} \sum_{j=0}^{m-1} c^T_{i,j}$ and $N_R \equiv \sum_{i=0}^{n-1} \sum_{j=0}^{m-1} c^R_{i,j}$, respectively. We define normalized

Tweet distributions by: $f^T_{i,j} \equiv (c^T_{i,j}/N_T)$ and $f^R_{i,j} \equiv (c^R_{i,j}/N_R)$, which represent the fraction of Tweets in the (i,j)th bin for the target and reference cases, respectively. The comparison between the two distributions is then done by calculating the difference in the Tweet fraction per bin: $\Delta f_{i,j} \equiv f^T_{i,j} - f^R_{i,j}$, which henceforth we refer to as the "relative distribution". This quantity can be understood as follows: bins with positive values of $\Delta f_{i,j}$ over represent the target Tweets while negative values under represent them, relative to the reference-Tweet distribution. The normalization ensures that the difference between the distributions will be identically zero (i.e., $\Delta f_{i,j} = 0$) for the hypothetical situation where the two distributions have the exact same shape but a different total number of Tweets.

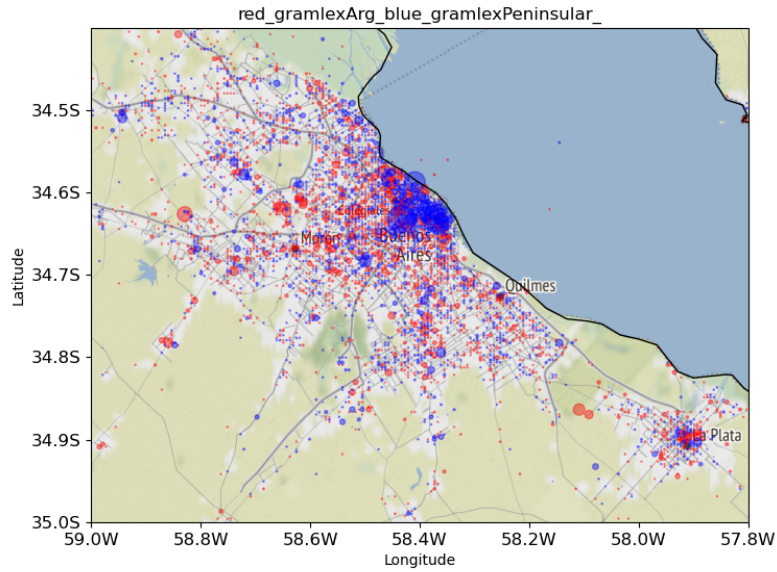

**Figure 3.** Relative distribution of ArgSp vs PenSp Tweets.

Figure 3 shows a plot of the *relative distribution* calculated from the distributions in Figure 2. Variations between the two distributions are now easily seen. This plot is similar to the geographic plots of the Tweet counts in that the magnitude of $\Delta f_{i,j}$ is represented by the size of the marker. However, since $\Delta f_{i,j}$ can be both positive or negative, a two-color representation was used, with red for the positive values and blue for the negative ones. As a result, locations where the *relative distributions* vary are highlighted by the contrast in color, and locations where the normalized distributions are the same are de-emphasized, since bins with $\Delta f_{i,j} = 0$ are represented by a marker of zero size and are thus not visible. Our formulation ensures that the sum of the distribution differences is identically zero: $\sum_{i=0}^{n-1} \sum_{j=0}^{m-1} \Delta f_{i,j} = 0$. This means that, overall, the sum of the positive (red) differences must be equal to the sum of negative (blue) differences. For the hypothetical situation of two distributions which have the same shape but different center locations, one can expect a bi-polar pattern of red and blue. It should be noted that in areas where Tweet counts are higher, the magnitudes of $\Delta f_{i,j}$ can also be expected to be larger. Therefore, interpretation of the calculation results must consider this effect.

## 2.3 Quantification of the variations

A key challenge of the methodology we present here is determination of the extent to which the patterns we measure can be reliably connected to an underlying phenomenon which drives the variation. A common approach, when evaluating a new technique, is to study a known phenomenon and confirm that the known behavior can be recovered. Although similar methods have been used to study language variation on large scales (see Eisenstein et al 2014, Nguyen et al 2016), to our knowledge, there are no comparable quantitative studies of language variation on a *micro-scale*. We therefore develop several metrics to compare the variations observed for different cases. The metrics are based on statistical analysis of the target and reference Tweet counts from each bin ($c^R_{i,j}$, $c^T_{i,j}$) taken as x,y coordinates of a scatter plot (henceforth "frequency-comparison plots") which we create for each measurement of $\Delta f_{i,j}$.

We start by testing for the likelihood that the *null* hypothesis is true, namely that the target and reference distributions have the *same* shape. This hypothesis can be expressed as $c^T_{i,j} = kc^R_{i,j}$, where $k$ is a proportionality constant that accounts for the difference in overall number of Tweets between the two distributions. This equation indicates that every point in the scatter plot lies exactly on a line with a slope equal to the ratio of the number of target and reference Tweets, i.e., $k = N^T/N^R$. For reference, this "exact-correlation" line is displayed on each scatter plot as a green dashed line. The extent to which the points do not lie on the line thus indicates how different the two distributions are. Points above and below the line correspond to areas that over- and under-represent target Tweets, respectively, relative to the reference Tweets, and are indicated by red and blue markers in the relative-distribution plots.

The degree of similarity between the two distributions is directly related to the degree of correlation between the target and reference Tweet counts. We therefore perform a linear regression analysis on the *frequency-comparison* data and calculate the Pearson correlation coefficient (see Wikipedia "Pearson correlation coefficient"). The best-fit line from this regression analysis is displayed on each plot as a solid red line. The deviations from exact correlation can be seen by two features in the *frequency-comparison* plot: 1) the spread of data around the best-fit line, and 2) the deviation of the best-fit line from the *exact-correlation* line. The Pearson coefficient provides a metric of the degree of spread of the data around the best-fit line. A higher correlation coefficient indicates two distributions which are more similar, while a lower coefficient indicates a greater variation between the two. A perfect correlation (equal distributions) results in a Pearson coefficient of of $r = 1$, while completely uncorrelated data (no similarity between the distributions) yields a coefficient of $r = 0$. A discrepancy between the slopes of the best-fit and exact-correlation lines, on the other hand, shows that the ratio between target and reference Tweet counts varies between bins with low and high Tweet counts.

A second metric for evaluating the two measures of deviation from exact correlation was also developed to complement the Pearson test. This metric was designed to give a more intuitive sense for the amount of spread or non-correlation in the frequency-comparison plots: for each point in the scatter plot the angle (relative to the x-axis) of the line connecting it to the origin is calculated. A histogram of the angles is then calculated by defining a set of bins of width 1 degree, over the range from 0 - 90 degrees, and counting the number of points in each bin. For exactly correlated data, the histogram would show a peak with zero width at the angle corresponding to the slope of the *exact-correlation* line. For data that is highly but not exactly correlated, the histogram then has a peak centered at the angle of the best-fit line and has a width which directly measures the degree of spread. Thus, the lower the degree of correlation (i.e., with more variation), the larger the width of the peak.

## 2.4 Comparing variations of different cases

The "significance" of a particular variation observation is evaluated by comparing test cases where variations are expected with "null" cases where variations are not expected. To make the comparison, it is important to understand and distinguish the potential sources that may contribute to the observed variation. The test cases can be broadly divided into two categories: "noise" and "signal". *Noise* refers to sources of variation that do not correlate to the particular phenomenon being studied, and include: 1) random statistical variations or "shot noise" such as observed in the classic coin-toss experiment [https://en.wikipedia.org/wiki/Shot_noise]; 2) inherent variability of human behavior, such as making mistakes or changing one's mind; 3) variations due to unanticipated or unresolvable correlations with location; and 4) skewing of patterns due to highly automated tweeters (i.e., "bots"). The first category of noise can be estimated by assuming it results from a random process. In random processes, the noise level associated with counting N individual events is equal to $\sqrt{N}$. The statistical noise in the target distribution can thus be described by $\Delta c^{T,stat}_{i,j} = k\sqrt{c^{R}_{i,j}}$. To display the level of shot noise that can be expected from this source, we added grey lines on either side of the exact-correlation line and separated from the line by the amount $\pm \Delta c^{T,stat}_{i,j}$.

The second and third categories of noise include associations that are not visible to an external observer between specific text in the Tweets and specific locations. For instance, the availability or accessibility of particular emojis or accented characters may depend on the device used for tweeting (e.g., iPhone vs Android-based) which could in turn correlate with social status and ultimately location within the city. The last category of noise, "bots", refers to automated sources which generate Tweets in ways that are not typical for people. These sources can generate Tweets in extreme quantities that may be highly repetitive in nature, and thus individual bots can easily induce strong features in the distribution which may or may not be consistent with the patterns produced by human tweeters. Identifying bots and evaluating their influence, however, is a complex task which we do not address here.

*Signal* variations, on the other hand, are those that can be correlated to identifiable sources, including: associations of a particular location with 1) a proper noun, such as a name; 2) a common noun expressing an activity or a social function, such as a sport or a job; and 3) tokens representing linguistic categories such as grammatical elements, language styles and language varieties. In general, strong signatures of correlation between the text of the Tweets and location can be expected from the first two categories because people tend to discuss topics and concepts that are relevant to who they are, where they are and what they are doing, i.e., on their social context. By contrast, finding differences in the distribution of linguistic categories such as language varieties or styles can be more difficult because, in the absence of physical borders, social interaction tends to homogenize language use, especially on small scales. In cities with a rich multicultural and sociolinguistic background, however, linguistic elements that mark an association with a specific group of people or a neighborhood can be expected (see Labov 2000, Smakman & Heinrich 2018, among others). As these associations are more complex and subtle than those associated with proper nouns, a highly sensitive technique is required to find and verify the variation patterns.

In the results section, we analyze several test cases selected to highlight variations expected from the *noise* and *signal* sources described above. For each case, a geographical plot of the relative distribution $\Delta f_{i,j}$ along with a frequency-comparison plot of $c^T_{i,j}$ vs $c^R_{i,j}$.

## 3. Results

In this section we show the results of our distribution calculations and comparisons. The primary goal is to demonstrate the ability to measure *micro-scale* patterns in language use and to show that these patterns can be correlated with highly localized phenomena that drive the variations. This demonstration is done by measuring patterns of variation from the noise and signal sources identified in section 2.4.

### 3.1 Variation due to noise sources

We start by examining "*null*" cases, for which we expect *no* significant difference in the shape of the target and reference distributions in order to provide a baseline for the level of noise that can be expected from the Twitter data (Fig. 4). For these null-cases we selected sets of target and reference tokens that do not express different meanings by themselves, but have the same or similar grammatical or social function. As a result, the pair of tokens sets can then be expected to be used in similar locations since they do not directly or indirectly mark a location. These sets can include tokens such as definite or indefinite articles, individual letters, emoticons or similar.

Case one (Fig. 4, top-left) compares Tweet distributions for the target token set {*las*} and reference token set {*los*}. The linguistic element pair (*las*, *los*) are the plural-feminine and plural-masculine versions of the definite article, 'the', and thus have the same function. As expected, the analysis does not show a strong spatial variation in the representation of either article. The distribution of red and blue points are evenly distributed over the extent of the calculation, with a few, larger variations tending to occur in the downtown area of CABA (upper-right quadrant of the graph). The larger variations downtown are expected due to the larger concentrations of people and hence larger overall density of Tweets, as seen in Figure 1. The frequency-comparison plot in Figure 4, top-right shows that the counts of target and reference Tweets in each bin are strongly correlated, with the points clustered tightly around the best-fit line. In addition, the slopes of the best-fit and exact-correlation lines match well, which, in addition to the other features of the two plots, indicates that the target and reference distributions are quite similar. Nevertheless, deviations of the points in the scatter plot from the exact-correlation line are noticeable and give us a measure of the level of variation to be expected from noise sources. This analysis shows that the observed variations are significantly larger than can be explained by pure statistical noise, and hence are most likely due to contributions from noise categories 2 - 4 described above.

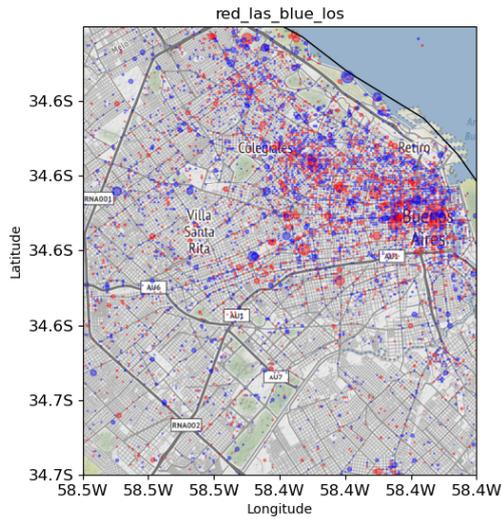
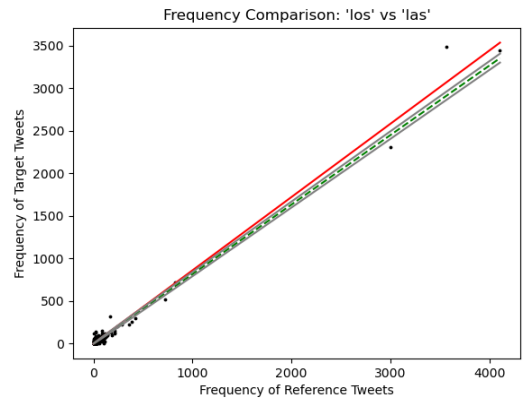

*r*(38998) = .992, *p* < .00,
Nr.Targ.= **33919**, Nr.Ref. = **41531**

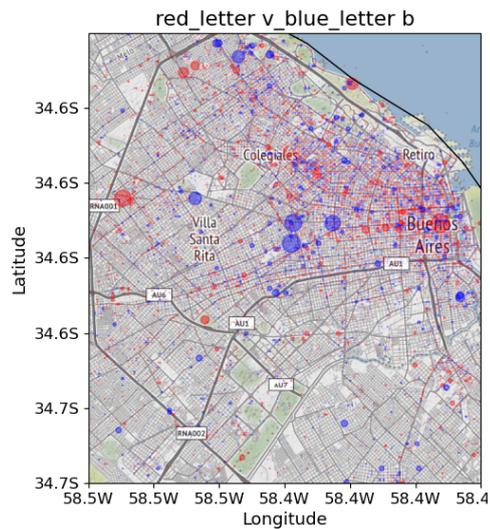
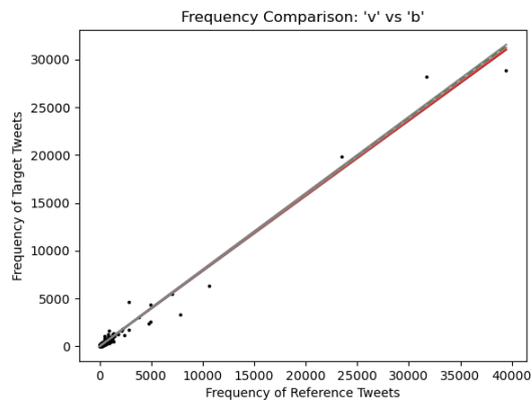

*r*(38998) = .990, *p* < .00,
Nr.Targ.= **316866**, Nr.Ref. = **398416**

**Figure 4:** Null cases. **Top-left:** Relative distribution plot for case 1 ("las" vs "los"). **Top-right:** Frequency plot for case 1. **Bottom-left and -right:** Relative distribution and Frequency plots for case 2 ("v" vs "b"), respectively.

For Case 2, we compared Tweets with two phonologically similar letters "b" and "v". Since the linguistic elements for this case are individual letters, the Tweet counts for each bin are significantly higher than for Case 1 where the linguistic elements were words. Nevertheless, both the geographical plot of the relative distribution and the scatter plot of the frequencies show a similar pattern to Case 1.

### 3.2 Signal variations from nouns with direct and indirect reference to location

The analysis of section 3.1 provides a baseline for how much variation to expect in cases where no correlation with location is foreseen. Definitive demonstration of our technique therefore requires observation of variations of a higher degree than found in the null cases. We therefore next examine cases where strong signal variations can be expected. For this task, we choose target and reference Tweets containing names of neighborhoods and nouns that express activities which, unlike grammatical markers, can be directly linked to particular locations. Tweets mentioning neighborhood names are especially suitable as there exists a unique connection to one specific location. For this test, we chose the names of two important neighborhoods in CABA, "La Boca" and "Palermo", for the target and reference Tweets, respectively. As shown in Figure 5-left, the difference in the distribution of the two names is striking. Tweets mentioning each of the names are highly localized to the centers of each respective neighborhood, and have very little geographic overlap with each other. The corresponding frequency-comparison plot confirms that target and reference Tweet counts correlate very poorly, and even appear to anti-correlate, which is a sign that the two distributions are entirely distinct.

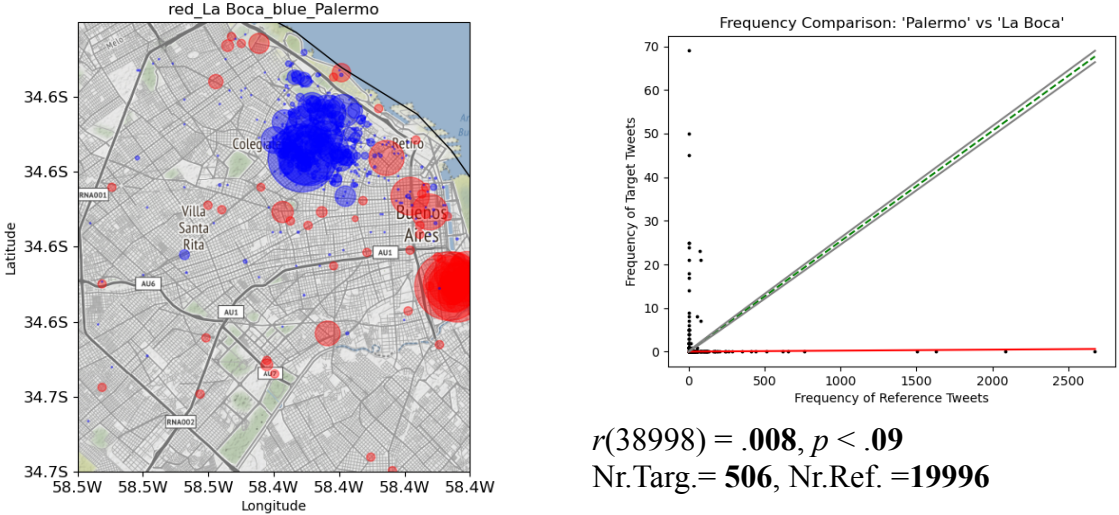

$r(38998) = .008$, $p < .09$
Nr.Targ.= **506**, Nr.Ref. =**19996**

Figure 5. Comparison of distributions of Tweets mentioning "La Boca" and "Palermo". **Left:** Geoplot. **Right:** Frequency comparison.

Tweets containing nouns that express recreational or work-related activities can also be connected with specific locations where these activities are performed. This connection is generally not as unique as for neighborhood names, but nevertheless should result in a strong and understandable pattern. Figure 6 shows an analysis of three such cases: Case 1 (Fig. 5 top-left) compares Tweets containing the increasing trend-line symbol 📈 with those containing the emoji 😊; Case 2 (Fig. 6 middle-left) compares Tweets mentioning *Tango* and *fútbol* ('soccer'); and Case 3 (Fig. 6 bottom-left) compares Tweets mentioning *plata* ('money') and *vacaciones* ('vacation').

All three cases show relative distributions (i.e., geographical plots of $\Delta f_{i,j}$) with patterns that are very distinctive compared to the null cases presented in section 3.1. The patterns are not as stark as in the case of "La Boca" and "Palermo" because the linguistic elements in these topic-related cases are not uniquely associated with particular locations. Nevertheless, the patterns can be understood by analysis of the locations associated with the most prominent red and blue areas in the plot of $\Delta f_{i,j}$. Local landmarks at these points, identified using Google Maps, were found to have a social function that correlated well with the respective topics represented by the linguistic elements of the query. Below we analyze each case separately.

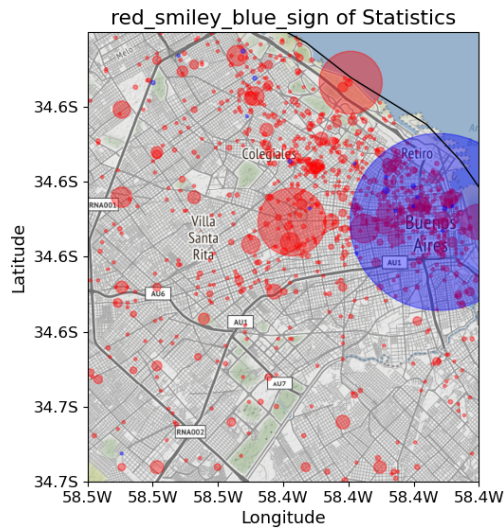
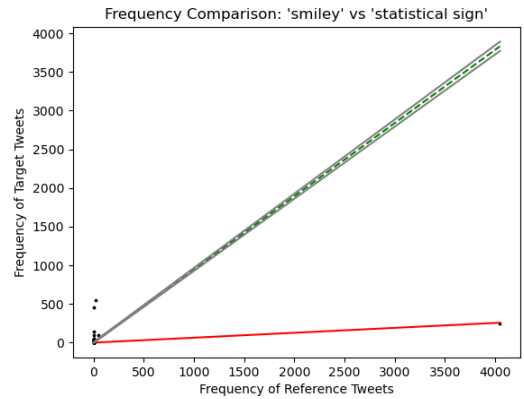

*r*(38998) = .**323**, *p* < .**00**
Nr.Targ.= **4178**, Nr.Ref. =**3956**

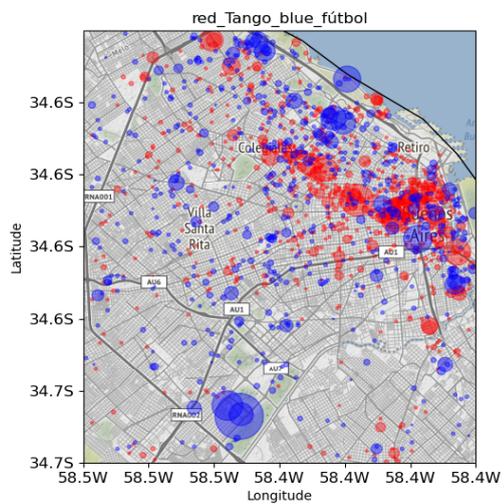
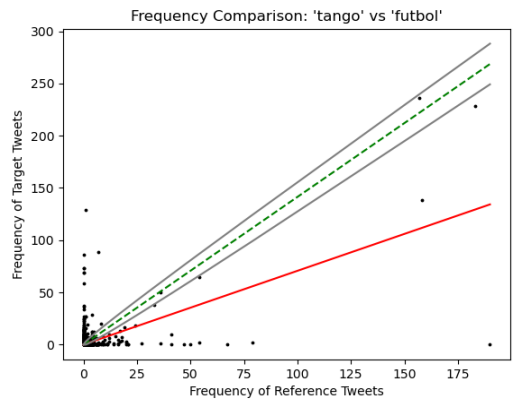

*r*(38998) = .**606**, *p* < .**00**
Nr.Targ.= **2852**, Nr.Ref. =**4032**

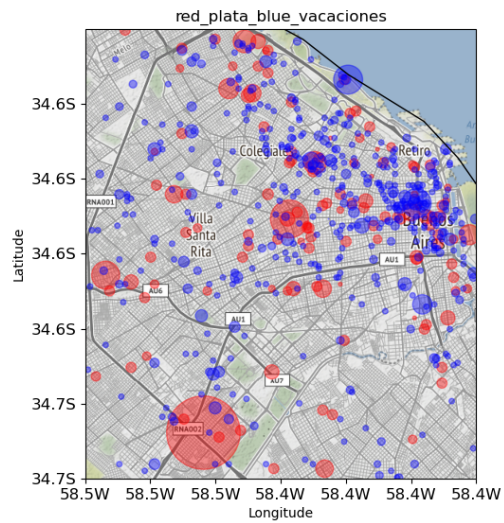
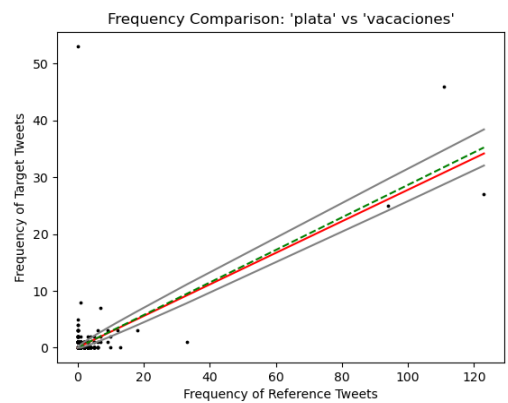

*r*(38998) = .**680**, *p* < .**00**
Nr.Targ.= **339**, Nr.Ref. =**1183**

**Figure 6.** Comparison of distribution of topics. **Top Row:** "😊" vs "📈". **Middle Row:** "tango" vs "fútbol". **Bottom Row:** "plata" vs "vacaciones".

Case 1: Tweets with the trend-line symbol, which is associated with stock prices and revenues, are almost exclusively localized to a single bin in the center of Buenos Aires near the *Bolsa de Comercio* stock exchange in Monserrat. Tweets with the emoji, which is generally used in informal settings to express a happy emotion, are distributed much more uniformly across the city. Case 2: The biggest concentration of the red area on the map coincides with tango clubs and dancing schools in CABA (especially in the neighborhoods of Monserrat and Colegiales). The biggest concentration of blue areas representing the topic *fútbol* coincides with soccer clubs, e.g., Defensores de Belgrano, Sports Lugano, Boca juniors. Case 3: The most represented places in which twitterers talk about *plata* are the *Centro Comercial Chilavert*, in the southwest of the city, shopping center in *Caballito* and *Bolsa de Comercio* in Monserrat. The 'vacation' topic, by contrast, is strongly represented at the airport Newbery Airport and the downtown region where airport taxis and shuttles can be hired.

To complement the distribution pattern analysis, Tweet-frequency correlation analyses were also done on the three cases (Fig. 6, top-, middle- & bottom-right). All show only a small amount of the data having a linear correlation and a larger fraction being anti-correlated, confirming the presence of a variation signal and that the two topics occupy distinct locations within the city. These results show that key words in the Tweets can effectively be used to correlate topics of the Tweets to respective locations associated to the topics by social activity. This information presents options for further use to analyse how social behavior and needs vary within the city.

### 3.3 Geolocation of dialectal variation

In the previous sections we demonstrated that for Tweets containing neighborhood names and activity-related nouns, clear patterns of variation can be observed which are significantly larger than those observed for the null cases. Here we look at the possibility of resolving variations in the use of a key socio-linguistic element: dialect. Studies of dialectal variation with location have previously been done on the scale of countries (see Hovy et al 2020). In this case, the geographic borders represent separations which restrict interactions between neighboring countries and are integrally involved in creating the linguistic variations. On much smaller scales such as within a city, however, social forces and fluid interactions between people tend to eliminate these variations. In addition, unlike names and activity-related nouns, tokens representing dialects do not directly encode particular locations, so variations are expected to be less obvious.

Buenos Aires, however, is a city that has a strong presence of Spanish dialects, in particular ArgSp and PenSp. These dialects play different roles in the society and may be associated with different groups of people, in part due to their different origins. PenSp is the *lingua franca* of the Spanish-speaking world, while ArgSp is a local dialect. As a result, these dialects can be expected to show patterns of use which are associated with their different roles. Tourists, for example, which come from many parts of the world, tend to communicate using PenSp. Tourist places can therefore be expected to have a larger representation of PenSp relative to ArgSp.

In order to test these assumptions, we analyze the relative distribution in the use of ArgSp relative to PenSp (Fig. 7). To select Tweets of each dialect, we checked for the presence of one or more of a set of words well-known in the linguistic literature to mark the two varieties (e.g., Conde 2011). The set of words included 2$^{nd}$-person singular verbal endings (e.g., PenSp *quieres* vs. ArgSp *querés* 'want'), which have grammatical roles and are known to be strongly connected to a particular variety. The geographical plot shows that the downtown is the area where PenSp. and Arg.Sp. differ most. This is illustrated by the large blue circles in the north eastern part of the city. The largest blue circle covers *La Plaza de Mayo* (right-middle of graph), the second largest covers a location on the avenue *Díaz Vélez* (middle of graph) and third largest, the Newbery airport in the northern part on the shore. *La Plaza de Mayo* is located in the financial district known as *microcentro*, within the *barrio* ("neighborhood") of *Monserrat*. The square is surrounded by significant historical monuments and important political departments (e.g., the seat of the President of Argentina), the Metropolitan Cathedral, the Buenos Aires City Hall, and the Bank of the Argentine Nation's headquarters. The left point on the avenue *Díaz Vélez* is surrounded by a hospital, a shopping center and the park *Parque Centenario*. Pen.Sp. is thus particularly concentrated in the downtown area of Buenos Aires associated with tourism and business, as expected. By contrast, red points that represent the locations where Arg.Sp. is more represented are more dispersed in the whole city. As Argentinian Spanish is the local dialect, this pattern is also expected.

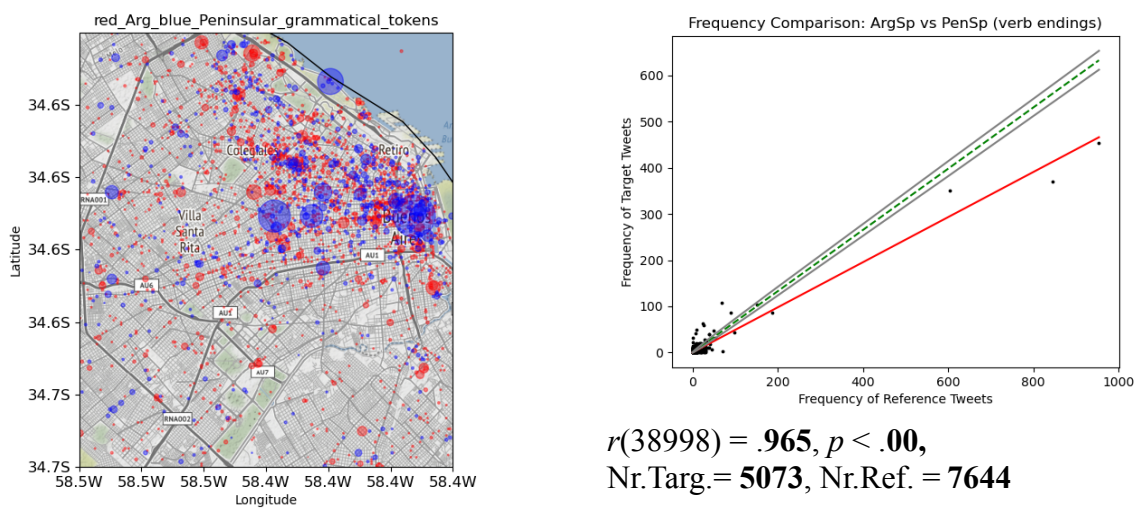

$r(38998) = .965, p < .00,$
Nr.Targ.= **5073**, Nr.Ref. = **7644**

**Figure 7. Left:** relative distribution of ArgSp vs PenSp grammatical endings. **Right:** comparison of Tweet counts per bin for relative distribution on the left.

3.4 Statistical comparison of Null, Location-related nouns and dialectal cases

In order to quantify the degree to which the variations observed in the dialectal data can be trusted, we compared the frequency analysis with those of the null case and one of the strong variation cases (see Fig. 8). In the left column, close-ups of the plots of target Tweet counts vs reference Tweet counts are shown. In the null case ('b' vs 'v'), the points are visually concentrated near the best fit and exact-correlation lines. By contrast, the points in the strong variation case ('tango' vs 'fútbol'), do not concentrate around the best-fit or exact-correlation lines at all. The dialectal variation case lies in between the two extremes. The angle histogram for the null case shows a Gaussian-like distribution of angles, with a standard deviation of 25.2 degrees, around the exact-correlation line, which confirms the visual interpretation of the frequency comparison plot. By contrast, the strong-variation case shows that there is essentially no preferential angle (standard deviation of 41.8 degrees). The dialectal variation case shows a nearly uniform distribution of angles, with a standard deviation of 39.1 degrees, that is significantly larger than that of the null case and only slightly smaller than that of the strong-variation case. These results confirm the expectation that the dialectal variations are less obvious than those associated with the activity-related nouns, but that the signal is still measurable.

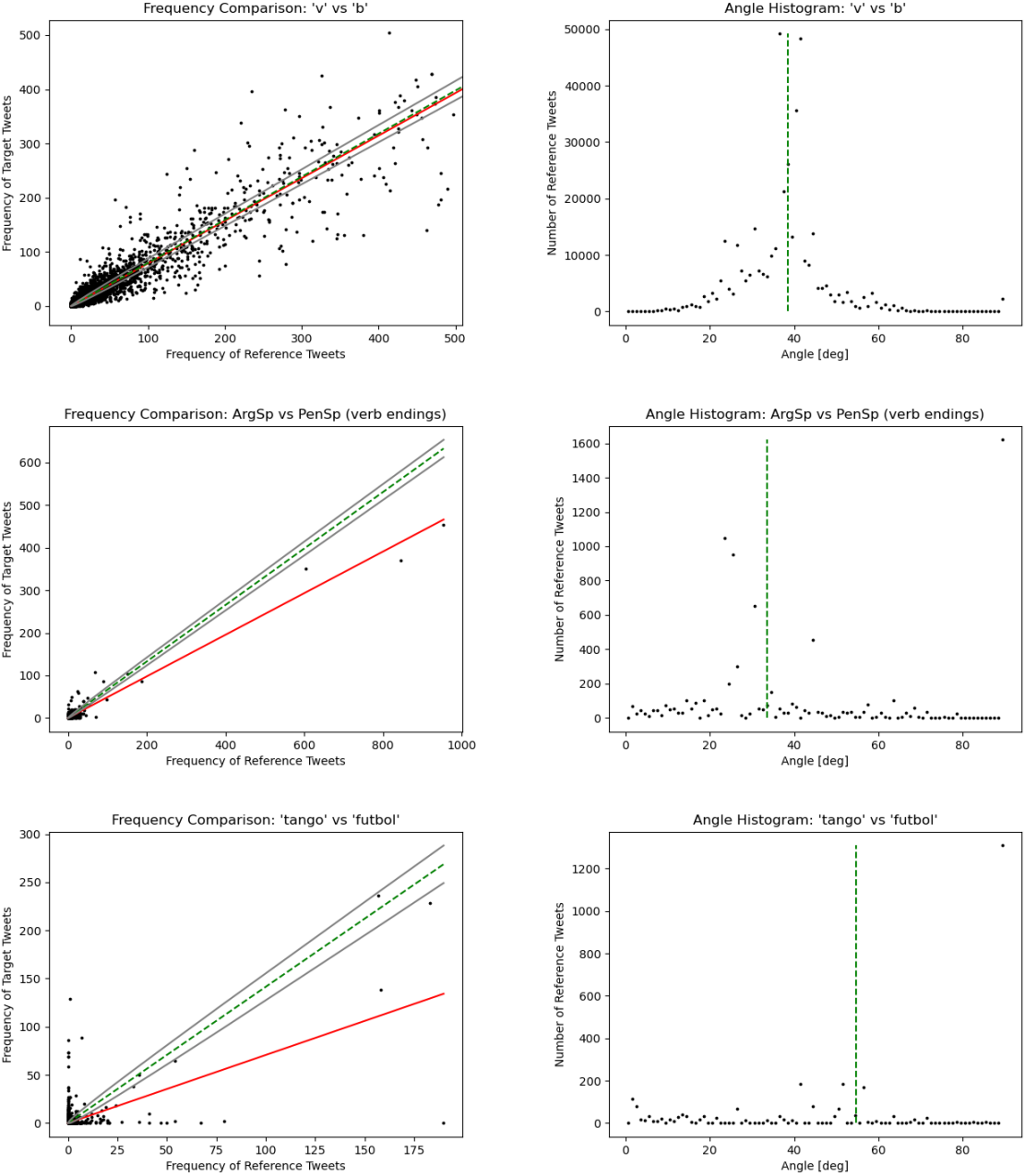

**Figure 8.** Comparison of frequency plots **(left)** and angle histograms **(right)** for null **(top)**, dialectal **(middle)** and activity-related noun **(bottom)** cases.

4.Conclusions and outlook

We have investigated the possibility of measuring variations in language use that occur on spatial scales that are smaller than the extent of a city, from neighborhoods down to single city blocks. Data for the investigation was sourced from Twitter, which is a social media platform providing text as well as complementary information such as precise geolocation information.

We find that variations in language use on these scales can be clearly observed, and provide evidence that these variations are real.

In order to evaluate the reliability of the patterns we observe, we first evaluated the base-line of variations to be expected in null cases when there is no underlying dependence and show that variations in the use of various language elements are more pronounced than for the null cases. The degree of variation depends strongly on the category of language element being studied. Four categories were studied: names of locations, common nouns related to a location, dialectal words and grammatical elements. Location names showed the most obvious patterns of variation, followed by common nouns, such as *tango*, related to a location via an activity. The dialectal variations, by contrast, are characterized by large variations in a few key locations and relatively lower level of variation elsewhere. Finally, the grammatical case shows a lack of significant variation, as evidenced by the high degree of correlation between the target and reference Tweet counts, justifying its use as the null case.

In this study, various types of language variation (e.g., use of location names and use of dialects) were selected by testing for presence of specific tokens of each type within the Tweets. As a result, the observed geolocation differences are unavoidably affected by the particular set of tokens chosen. A future possibility is to complement this approach by using other techniques such as machine learning to identify Tweets of particular dialects as has been done with dialectal data on larger geographic scales (see Hovy et al 2020, among others).

Our work opens up opportunities for many applications with potential for social, economic and scientific benefit. These include determining the locations associated with specific topics such as dancing or sports to help guide placement of businesses and resources as well as analyzing variations in use of language varieties for linguistic and social sciences as well as in order to tailor to the needs of people from various cultural and social backgrounds, leading in the direction of the development of smart cities (Milusheva et al. 2021).